\documentclass[10pt,twocolumn,letterpaper]{article}
\usepackage{cvpr}
\usepackage{times}
\usepackage{epsfig}
\usepackage{graphicx}
\usepackage{amsmath}
\usepackage{amssymb}
\usepackage[nolist]{acronym}

\cvprfinalcopy % *** Uncomment this line for the final submission

\def\cvprPaperID{5798} % *** Enter the CVPR Paper ID here

\begin{document}
\begin{acronym}
    \newacro{BB}{Bounding Box}
    \newacro{COCO}{Common Objects In Context}
	\newacro{CNN}{Convolutional Neural Network}
    \newacro{ILSVRC}{Imagenet's Large Scale Visual Recognition Challenge}
	\newacro{IoU}{Intersection over Union}
	\newacro{mAP}{mean Average Precision}
	\newacro{MDP}{Markov decision process}
	\newacro{RoI}{Region of Interest}
	\newacro{RPN}{Region Proposal Network}
	\newacro{SSD}{Single Shot Detector}
	\newacro{OF}{Optical Flow}
    \newacro{FC}{Fully-Connected}
    \newacro{NHU}{Number of Hidden Units}
    \newacro{MLP}{Multi-Layer Perceptron}
    \newacro{RNN}{Recurrent Neural Network}
    \newacro{LSTM}{Long short-term memory}
    \newacro{TSN}{Temporal Segment Network}
    \newacro{R-CNN}{Region-Convolutional Neural Network}
    \newacro{MLD}{Multi-label dataset}
    \newacro{SLD}{Single-label dataset}
    \newacro{BPTT}{Backpropagation Through Time}
    \newacro{SVM}{Support Vector Machines}
    \newacro{RF}{Random Forests}
    \newacro{AVA}{Atomic Visual Actions}
    \newacro{GPU}{Graphics Processing Unit}
    \newacro{UCF}{University of Central Florida}
    \newacro{BB}{Bounding Box}
    \newacro{AP}{Average Precision}
    \newacro{iDT}{improved Dense Trajectories}
    \newacro{FPS}{Frames Per Second}
    \newacro{GBB}{Gaussian Background Blur}
\end{acronym}

%%%%%%%%% TITLE
\title{Attention Filtering for Multi-person Spatiotemporal Action Detection on Deep Two-Stream CNN Architectures}
\author{Jo\~{a}o Antunes\\
Carnegie Mellon University\\
Instituto Superior T\'{e}cnico\\
{\tt\small joaoa@andrew.cmu.edu}
\and
Pedro Abreu\\
Institudo Superior T\'{e}cnico\\
{\tt\small pedro\_abreu95@hotmail.com}
\and
Alexandre Bernardino\\
Institudo Superior T\'{e}cnico\\
{\tt\small alex@isr.tecnico.ulisboa.pt}
\and
Asim Smailagic\\
Carnegie Mellon University\\
{\tt\small asim@cs.cmu.edu}
\and
Daniel Siewiorek\\
Carnegie Mellon University\\
{\tt\small dps@cs.cmu.edu}
}

% \author{
% Jo\~{a}o Antunes\\
% Carnegie Mellon University\\
% Instituto Superior T\'{e}cnico\\
% {\tt\small joaoa@andrew.cmu.edu}
%     \and
% Pedro Abreu\\
% Instituto Superior T\'{e}cnico\\
% {\tt\small pedro$\_$abreu95@hotmail.com}
%     \and
% Alexandre Bernardino\\
% Instituto Superior T\'{e}cnico\\
% {\tt\small alex@isr.tecnico.ulisboa.pt}
%     \and
% Asim Smailagic\\
% Carnegie Mellon University\\
% {\tt\small asim@cs.cmu.edu}
%     \and
% Daniel Siewiorek\\
% Carnegie Mellon University\\
% {\tt\small dps@cs.cmu.edu}
    
% }

\maketitle

%\thispagestyle{empty}

%%%%%%%%% ABSTRACT
%\input{0.abstract}
\begin{abstract}
Action detection and recognition tasks have been the target of much focus in the computer vision community due to their many applications, namely, security, robotics and recommendation systems.
Recently, datasets like AVA, provide multi-person, multi-label, spatiotemporal action detection and recognition challenges. Being unable to discern which portions of the input to use for classification is a limitation of two-stream CNN approaches, once the vision task involves several people with several labels. We address this limitation and improve the state-of-the-art performance of two-stream CNNs. In this paper we present four contributions: our fovea attention filtering that highlights targets for classification without discarding background; a generalized binary loss function designed for the AVA dataset; miniAVA, a partition of AVA that maintains temporal continuity and class distribution with only one tenth of the dataset size; and ablation studies on alternative attention filters. Our method, using fovea attention filtering and our generalized binary loss, achieves a relative video mAP improvement of $20\%$ over the two-stream baseline in AVA, and is competitive with the state-of-the-art in the UCF101-24. We also show a relative video mAP improvement of $12.6\%$ when using our generalized binary loss over the standard sum-of-sigmoids.

\end{abstract}
%%%%%%%%% BODY TEXT
\section{Introduction}
When ImageNet \cite{Russakovsky2015} appeared, image classification performance increased dramatically. The features learned from data greatly outperformed hand-crafted features and the methods found in the state-of-the-art of that task all converged to \ac{CNN}s. However, such a dramatic paradigm shift has yet to be discovered when video understanding tasks are researched \cite{tran2018closer}. 3D-CNN models are currently the approach with the best performance, but the gap between data-driven feature extraction and hand-crafted features like iDT \cite{idt} is surmountable. Also, the performance of 2D-CNN models applied to still video frames shows competitive performance with 3D-CNN's, despite not incorporating any type of temporal information modelling. Furthermore, 2D-CNN models have fewer parameters to train, making them computationally more tractable. Architectures such as two-stream 2D-CNN's, that take RGB and Optical Flow as input, offer a good balance between having temporal information encoding on optical flow with context on the RGB frames, but still not requiring the immense computational power that 3D-CNN models need.

Recently, datasets such as the AVA dataset \cite{ava} or the UCF101-24 \cite{soomro2012ucf101} have increased the complexity of the video understanding task by providing multiple persons and labels per video instance. Since for each video there may be an arbitrary number of persons performing an arbitrary number of actions, the models used in those datasets are required to be able to identify each person on the video, and only then classify the actions being depicted. The difficulty in data region selection is a limitation found on state-of-the-art two-stream \ac{CNN} approaches. In this work, we address this limitation. We propose an attention filtering mechanism to be used with CNN based classifiers. Using attention filtering, we can ensure that the classification models we employ use more detailed information in the location of the target rather than in the surrounding background. This is particularly useful in cases where different actors performed different actions at the same time, as is the case of the AVA dataset.
%In a multi-person classification problem there are crucial sections of information in the input data, like the person whose actions are being classified, but also data that is secondary to the classification task, like people in the frame who are not the target being classified or background.

\subsection{Contributions}
There are four main contributions in this work:
\begin{itemize}
    \item Fovea attention filtering, a filtering method mimicking human vision. This technique can be applied with any classification approach, and resulted in significant performance increases on our tests with a state-of-the-art two-stream \ac{CNN} (20\% relative performance increase in video mAP on the AVA dataset)
    \item A custom loss function to be used with the popular AVA dataset that takes into account this dataset's unique labelling structure
    \item miniAVA, a partition of the AVA dataset that maintains temporal continuity, keeps the same approximate distribution of the data, while being only 10\% of the full dataset
    \item Ablation studies on alternative attention filtering approaches
\end{itemize}
%All of our code, models, results and the miniAVA data are available online at \url{https://github.com/pedro-abreu/twostream-attention}.

%Due to computational limitations there were many compromises we had to do. Namely on model complexity and on dataset size. Still, since all methods we tested were studied on the same conditions we believe the study presented here to be a solid proof of our method's merits. Throughout the paper we mention every time a decision was made due to computational limitations, and what could be done more thoroughly where possible.
\subsection{Paper Outline}
The rest of this paper is organized as follows: in Section \ref{sec:sota} we present an overview of the previous relevant work. In Section \ref{sec:ava} we describe the unique characteristic of the AVA dataset that guided several of our design choices. Section \ref{sec:methodology} describes our fovea attention filtering approach and our generalized binary loss function. In Section \ref{sec:experiments} we describe the experiments we ran to validate our approach as well as its implementations details. We present and discuss our results in Section \ref{sec:results} and conclude this work in Section \ref{sec:conclusions} with some final thoughts and future work considerations.

\section{Previous Work}
\label{sec:sota}
We can separate approaches to video classification in two categories: before \ac{CNN}s and after \ac{CNN}s. Before \ac{CNN}s, most approaches were heavily focused on feature representation and feature encoding. These are called \textit{handcrafted} since the features were often computed via an interpretable, analytical method. These approaches, of which \ac{iDT} \cite{idt} is a good example, can be broken down into the following three steps: i) high-dimensional visual features that describe a region of the video are extracted and linked across time; ii) the extracted features get encoded into a fixed-sized video description vector via BoVW \cite{Csurka04visualcategorization}, LLC  \cite{BMVC.25.76}, FV \cite{fisherencoding}\cite{Reynolds2009GaussianMM} or VLAD  \cite{5540039}; iii) a classifier, often \ac{SVM} or \ac{RF} is trained on the encodings.

After the performance increase achieved by \ac{CNN}s in the domain of image recognition \cite{Russakovsky2015}, it was believed that the advances in performance would also show improvement action recognition  \cite{DBLP:journals/corr/abs-1711-09577}. 
However, many handcrafted methods still outperformed \ac{CNN}-based methods regardless of the growing interest in the use of convolutions for action recognition over temporal stacks \cite{6165309}. Many network-based approaches  \cite{DBLP:journals/corr/TranBFTP14}\cite{DBLP:journals/corr/CarreiraZ17} still use an averaging with \ac{iDT} predictions in order to improve performance. Early network-based approaches \cite{karpathy} tried using single 2D \ac{CNN}s and their results were worse than traditional methods for two reasons: the learned features did not capture either motion and long temporal context and the UCF101  \cite{soomro2012ucf101} dataset was not large or varied enough for the networks to learn such features. 
After data-driven approaches became more powerful, the use of \ac{CNN} architectures in action recognition quickly became the state of the art  \cite{DBLP:journals/corr/TranBFTP14}. More recent action recognition approaches normally differ in how they employ temporal information  \cite{DBLP:journals/corr/CarreiraZ17}. We describe five main categories of approaches in this sense.

% \begin{figure*}
%   \includegraphics[width=\textwidth]{images/recurrent_theme_high.png}
%   \caption{Examples of typical action recognition approaches with \ac{CNN}s as shown in \cite{DBLP:journals/corr/CarreiraZ17}. Approaches d) and e) tend to be computationally intensive while a) and b) two tend to lack explicit motion features like Optical Flow. The Two-Stream approach presents an interesting trade-off of providing competitive results with the state-of-the-art 3D-\ac{CNN} approaches while being much less computationally demanding. See Sec. \ref{sec:sota} for more details. Image taken from \cite{DBLP:journals/corr/CarreiraZ17} with permission.}
% \label{fig:recurrent_theme}
% \end{figure*}
%It should be noted that the rise of \ac{CNN} approaches brought yet another design choice: there is also the option of end-to-end training vs feature extraction and classification. End-to-end training often involves more complex models that need fine-tuning and are prone to issues such as overfitting. On the other hand, feature extraction and classification involves a 2-step procedure that may be inefficient to train and test. We explain roughly the idea of the approaches in Fig. \ref{fig:recurrent_theme}:
\ac{LSTM} networks, (\textit{e.g} Beyond Short Snippets  \cite{DBLP:journals/corr/NgHVVMT15}, LRCN  \cite{DBLP:journals/corr/DonahueHGRVSD14}) model sequences \cite {DBLP:journals/corr/abs-1803-01271} of trained features maps extracted with \ac{CNN}s to capture long temporal information. Recently, SSN  \cite{DBLP:journals/corr/abs-1712-05080} have shown good results on long untrimmed videos. These approaches have the largest temporal receptive fields. 

3D-\ac{CNN}s, (\textit{e.g} C3D \cite{DBLP:journals/corr/TranBFTP14} \cite{DBLP:journals/corr/XuDS17}) built on earlier single-stream 2D approaches  \cite{karpathy}, but instead used 3D convolutions on video volumes.

Two-Stream \ac{CNN}, (\textit{e.g} Two-Stream \cite{simonyan}, Two-Stream Fusion  \cite{feichtenhofer-fusion}, TSN \cite{tsn}) introduce the use of an additional 2D-\ac{CNN} stream trained on explicit motion features like optical flow to try and capture short temporal context. Many aspects of this original approach have been further augmented (\textit{e.g} ActionVLAD  \cite{DBLP:journals/corr/GirdharRGSR17}, HiddenTwoStream  \cite{DBLP:journals/corr/ZhuLNH17a}). This approach presents an interesting trade-off in that they are capable of achieving competitive results with methods requiring far more computational resources like 3D-\ac{CNN}s.

Finally, Fused Two-Streams, and Two-Stream 3D-ConvNet, (\textit{e.g} I3D  \cite{DBLP:journals/corr/CarreiraZ17}, S3D  \cite{DBLP:journals/corr/abs-1712-04851},  R(2+1)D\cite{DBLP:journals/corr/abs-1711-11248}) are currently the best performing approaches. They extend two-stream approaches by in the time dimension as well, either through 3-D convolutions or by alternating between spatial 2-D convolutions with 1-D temporal convolutions. As such, these approaches tend to be more computationally intense than all others.

Two fundamental insights to be extracted when looking at these approaches is that allowing the network to learn temporal features is fundamental for good performance in video classification; and that incorporating explicit motion features like optical flow is a benefit to be found even in 3-D \ac{CNN}s.

In this work, we test our approaches for attention filtering using a state-of-the-art Two-Stream network. We were unable to use more complex 3D-\ac{CNN} models because of computational constraints: all experiments ran for this paper were computed on a machine with a single NVIDIA 1080Ti, with 32gb of RAM. For comparison, in \cite{DBLP:journals/corr/CarreiraZ17} the computations are run on machines with 32 or 64 GPUs. Nevertheless, we believe our approach could be used in more complex models. When compared using state-of-the-art networks, our attention filters resulted in a performance increase in all instances tested.

\section{The AVA dataset}
\label{sec:ava}
%We tested our attention filtering approach on both the AVA\cite{ava} dataset and the UCF101-24\cite{soomro2012ucf101}. However, the AVA dataset contains several specific design choices that make it unique, and explaining our method without referring the dataset characteristics that motivated our design choices would make it hard to properly convey the thinking behind the decision. So, in this section, the AVA set is described. Since the UCF101-24 dataset doesn't have anything that needs explaining, we will refer to it in Section \ref{sec:experiments} and not here.

The AVA dataset is composed of densely annotated 15-minute videos. The annotations are person-centric with a sampling frequency of 1Hz, and each annotation corresponds to a 3-second video snippet, with the label corresponding to the frame in the center, called the keyframe, of that snippet (this implies that two adjacent annotations have an overlap between them of 1.5 seconds).

 \begin{figure}[ht]
  \centering
  \includegraphics[width=.8\linewidth]{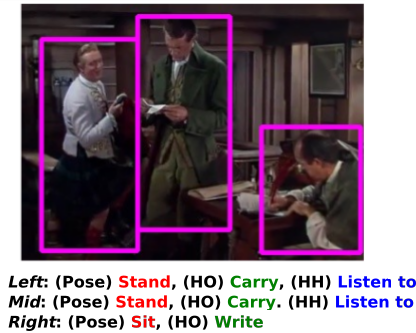} 
      \caption{Examples of AVA \cite{ava} labeling. \textit{HO} stands for human-object actions and \textit{HH} stands for human-human actions.}
  \label{fig:ava_labels}
\end{figure}

The dataset has a unique multi-label structure as shown in Fig. \ref{fig:ava_labels}. For each person, in each frame, a bounding box is provided as well as up to seven labels. Each bounding box in a frame must have 1 pose label like \textit{stand} or \textit{walk} (from mutually exclusive $C_P$ of them), 0-3 human-human labels (from non-mutually exclusive $C_H$ of them) like \textit{talk to} or \textit{hit} and 0-3 human-object labels (from non-mutually exclusive $C_O$ of them) like \textit{hold object} or \textit{watch (e.g TV)}. The total number of classes is $C = C_P + C_H + C_O$.

In AVA, the action class distribution is heavily imbalanced, roughly following Zipf's Law (see \cite{ava} for more details.) This imbalance itself is a challenge given that some classes will be severely under-represented (the most prevalent classes have over $10^5$ instances while the least common have less than 20).

Finally, to give the reader an approximate notion of how challenging this dataset is, we highlight the performance that the baseline model provided in \cite{ava} with this dataset achieved. The baseline model is a very deep, 3D-CNN that is a fusion of I3D \cite{DBLP:journals/corr/CarreiraZ17} and Faster RCNN \cite{DBLP:journals/corr/RenHG015}. The performance this model obtained in JHMDB \cite{jhmdb} and UCF101-24 \cite{soomro2012ucf101} beats other state-of-the-art methods \cite{ava}, achieving 78.6\% and 59.9\% video mAP at 0.5 IoU, respectively. In the AVA dataset, this model scored 12.3\% video mAP at 0.5 IoU.
\section{Proposed Methodology}
\label{sec:methodology}
The AVA dataset allows methods to be tested on a variety of challenges simultaneously, namely, action recognition but also action segmentation in both time and space. While we are interested in approaching the multi-person multi-label part of this dataset, we do not address the segmentation portion of the dataset. As such, our approach focuses on correctly classifying the actions being depicted by every person in a video, but not in localizing the actor in space and time. (\textit{i.e.} we focus on the classification part of the problem, and assume that the actor detection is a problem solved \textit{a priori}. We argue that using bounding box ground truth information doesn't reduce the difficulty of this dataset in a meaningful way by referring to the ablation study found in \cite{ava}. In Table 5 of  \cite{ava}, the authors report that comparing UCF101-24 and AVA, the performance decrease in actor detection (\textit{i.e.}: locating the person) is quite small (84.8\% to 75.3\% frame-mAP) when compared to the performance dip in action detection (76.3\%  to 15.6\% frame-mAP).

The rest of this section is ordered as follows: in Section \ref{subsec:attfilt} we describe the attention filtering methods studied in this work, Gaussian Background Blur (GBB), Crop filtering and Fovea filtering. In Section \ref{subsec:twostream} we describe the architecture we use for classification. We describe our custom Generalized Binary loss function in Section \ref{subsec:loss}, and finish the explanation of our methodology in Section \ref{subsec:ucf} with considerations on how to apply our method to the UCF101-24 dataset.

\subsection{Attention Filtering}
\label{subsec:attfilt}
Attention filtering contributes to improve classification performance by using more detailed information in the location of the target rather than in the surrounding background. As such, the features correlated to the action performed by the target person will have a stronger role in the classification than the "distracting features" of the other actors. This is particularly useful in cases where different actors performed different actions at the same time, as is the case of the AVA dataset. Two alternative approaches are possible: the first, which we refer to as Crop filter, would be to crop the area outside that region and the second, which we refer to as GBB (Gaussian Background Blur) would be to simply apply a Gaussian blur with a Gaussian kernel to everything outside that region. %The Crop filter approach is in effect what is happening when a network following the architecture defined by Faster R-CNN\cite{DBLP:journals/corr/RenHG015} is used, since the feature maps passed to the classifier layers of the network are cropped around the RoI chosen by the region proposal network.

Despite the goal of the attention filters, the Crop filter, having no background context at all, might hinder the classifier (i.e an abundance of blue might help a classifier guess the \textit{swim} action). Additionally, for both the GBB and the Crop filter, regions with large contrasts exist (i.e in the crop filter there is a sharp transition from the attention region to black and in the GBB there is a sharp transition from the relevant region to a blurred region).

For these edge artifacts we use an artificial foveal vision filter from  \cite{rui} inspired by human foveal vision  \cite{Traver2010ARO}\cite{laplacianpyramid}. This provides a smooth blur transition between the bounding box and the background. Firstly, a Gaussian pyramid is built with increasing levels of blur. The image $g_{k+1}$ can be obtained from $g_k$ via convolution with 2D isotopic Gaussian filter kernels with progressively higher $\sigma_k = 2^{k-1}\sigma_1$ standard deviations for each $k$th level. Next, a Laplacian pyramid \cite{laplacianpyramid} is computed from the difference between adjacent Gaussian levels. Finally, exponential weighted kernels are multiplied by each level of the Laplacian pyramid to emulate a smooth fovea:
\begin{equation}
    k(u,v,f_{kx},f_{ky}) = e^{-\left(\frac{(u-u_0)^2}{2f_{kx}^2} + \frac{(v-v_0)^2}{2f_{ky} ^2}\right)}, \qquad 0 \leq k \leq K
    \label{eq:elipticalfovea}
\end{equation}
where $f_{0x} = \frac{1}{2} w$, $f_{0y} = \frac{1}{2}h$, and $f_{kx} = 2^k f_{0x}$, $f_{ky} = 2^k f_{0y}$ is used to define the fovea intensity at the $k$-th level and the fovea is centered at $u_o = x + w/2$, $v_o = y + h/2$, given $(x,y,h,w)$ where $x,y$ are top-left corner coordinates and $h,w$ are the height and width of the bounding box.

\begin{figure*}[ht]
  \centering
  \includegraphics[width=\textwidth]{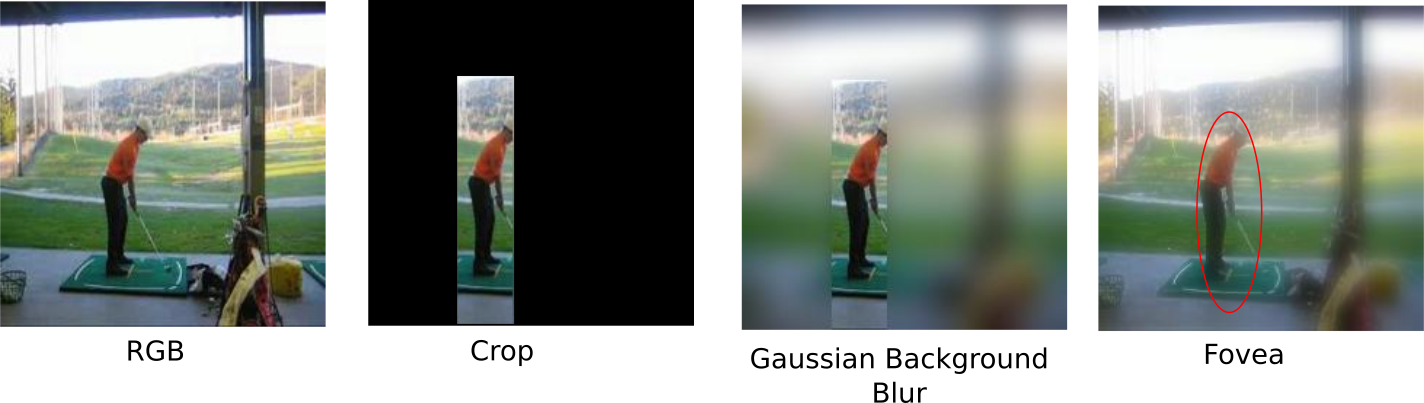}
  \caption{Comparison of the different attention filters. We highlight how our Fovea attention filter induces less artifacts in the image when compared with \ac{GBB} and Crop filtering.}
  \label{fig:attention filters}
\end{figure*}

In Figure \ref{fig:attention filters} we show an example of the attention filters used in this work, highlighting the difference in the abrupt transition boundaries introduced by the approaches of \ac{GBB} and Crop filtering versus the smooth transition in our Fovea attention filter. %We believe that these vertical edges will be relevant artifacts when doing classification with a CNN since it has been shown that the early layers on a CNN focus mostly on edge detection\cite{zeiler2014visualizing}. The results we obtained in our tests corroborate this hypothesis.
\subsection{Two-Stream \ac{CNN}}
\label{subsec:twostream}
\begin{figure*}[ht]
  \centering
  \includegraphics[width=\textwidth]{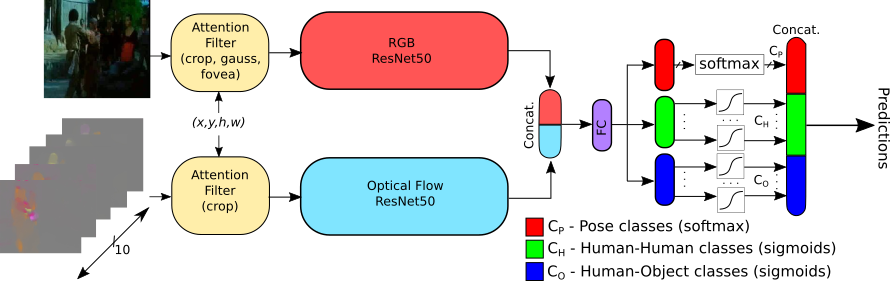}
  \caption{Our proposed two-stream architecture for the AVA dataset. After training each stream (RGB and OF) we fuse then with concatenation (see Section \ref{subsec:twostream}) and apply our Generalized Binary Loss on the output.  More information on our Generalized Binary Loss in Sec. \ref{subsec:loss}}
  \label{fig:twostreammodel}
\end{figure*}

We propose a baseline two-stream \ac{CNN} architecture motivated by promising results achieved with these networks \cite{feichtenhofer-fusion}\cite{feichtenhofer-multi}, as shown in Fig. \ref{fig:twostreammodel}. We use a ResNet50 architecture for each stream because this architecture demonstrated to be successful \cite{feichtenhofer-resnet} for these tasks, as well as having available pre-trained weights from  \cite{feichtenhofer-fusion} trained on UCF101 \cite{soomro2012ucf101}. The input is a single frame for the RGB stream and an optical flow stack for the OF stream. The output layers take into account the unique labelling structure of the dataset where for the pose classes ($C_P$) a Softmax is used, enforcing mutual exclusion between labels. For the human-human interaction ($C_H$) and human-object ($C_O$) interaction classes, a sum of sigmoids loss function is used, which is then thresholded. The top 3 values that are above the threshold are chosen as classes. Each network stream (RGB and OF) is individually fine-tuned for our task.
We fuse these networks using a concatenation fusion of their last Fully Connected (FC) layers and then train a FC layer so that spatiotemporal features can be learned. Since retraining both networks together would be too computationally heavy we load their individually fine-tuned weights, and freeze them (\textit{i.e} do not update them in backpropagation) so we only train the desired FC spatiotemporal filters.

\subsection{Generalized Binary Loss}
\label{subsec:loss}
The baseline method for AVA \cite{ava} has an an output layer of $C$ (number of classes) independent sigmoid activation functions. Note, however, that this does not truly reflect the label structure of the AVA \cite{ava} dataset which is not entirely mutually exclusive but also not entirely multi-label.  In the AVA dataset there are three distinct types of classes, each describing a type of action: pose classes (\textit{e.g.} standing), human-human interaction classes (\textit{e.g.} hug), and human-object interaction classes (\textit{e.g.} hold (an object)). While the pose classes are mutually exclusive (\textit{i.e.} each data sample has exactly one pose class in its label), the human-object and human-human interaction classes are not mutually exclusive, and any number of zero to three of each type can be found in the label of each data sample.

As such, in our architecture, we use our custom loss function: Generalized Binary Loss, to address this heterogeneity by having three separate output layers. The layer corresponding to the pose classes has size $C_P$ with a Softmax activation function. The remaining two layers have a sum-of-sigmoids activation function, where the number of sigmoids is $C_H$ for the human-human interaction classes and  $C_O$ for the human-object interaction classes.

This implies that the loss function cannot simply be categorical cross-entropy or binary cross-entropy \cite{deeplearning}. Therefore the architecture must minimize a global loss which is the sum of the losses of each output layer and which can be expressed as follows:
\begingroup\makeatletter\def\f@size{7}\check@mathfonts
\begin{multline}
     \mathcal{L}_{GB}= \overbrace{-\log\left(\frac{e^{s_p}}{\sum_j^{C_P} e^{s_j}}\right)}^{pose\ classes\ loss} + \overbrace{\sum_j^{C_H} \left(- \sum_{i=0}^{1} t_{j,i} \log(\sigma(s_{j,i})) \right)}^{human-human\ classes\ loss} + \\ \overbrace{\sum_j^{C_O} \left(- \sum_{i=0}^{1} t_{j,i} \log(\sigma(s_{j,i})) \right)}^{human-object\ classes\ loss}
    \label{eq:customloss}
\end{multline}
\endgroup
where the $s$ vectors are the predictions of each layer in equation \ref{eq:customloss} (different for each component of the sum), $s_p$ is the only positive label in the softmax layer, and $\sigma$ is the sigmoid function, $j$ is iterating over the classes, $i$ is iterating over the possible labels for each class ($1$ or $0$) and $t$ is the ground truth class.
Note that this backpropagation is possible as the encoding of the target labels is done as a binary vector which can then be partitioned into smaller vectors for each output layer. A representation of how this loss is incorporated in our architecture is shown in Fig. \ref{fig:twostreammodel}.

\subsection{UCF101-24}
\label{subsec:ucf}
In the UCF101-24 dataset, for each input video there is exactly one label identifying the action that is represented in it. When there are multiple people in the video, several bounding boxes are provided, but the label is the same for all the bounding boxes. As such, to use the architecture we show in Fig. \ref{fig:twostreammodel} on this dataset we change the output layers to a Softmax layer (since every class will be mutually exclusive with all other classes).

\section{Experiments}
\label{sec:experiments}
We test our models on two datasets, the UCF101-24 and the AVA datset. Although we could process the whole UCF101-24 in the computer we had available, due to computational limitations, we were unable to run the tests on the full AVA dataset, so we were forced to create a subset of this dataset. For simplicity, we refer to it as miniAVA.

\subsection{miniAVA}
The AVA dataset was too large for us to process with the computational resources we had available. In Table \ref{tab:datasets} we show the two datasets used in this work. We note that the AVA dataset is around 17 times larger than the UCF101-24. Furthermore, the multi-label characteristic of AVA makes it more computationally intensive during training. We made a partition of AVA, that, for simplicity, we call miniAVA.
\begin{table}[]
\centering
\begin{tabular}{ccc}
\hline
Dataset              & AVA      & UCF101-24 \\ \hline
\#Videos             & 430      & 3194      \\
Avg. Video Duration  & 15 min   & 7 sec     \\
\#Classes            & 80       & 24        \\
Labels per Video     & Variable & 1         \\
Total Video Duration & 6450 min & 373 min   \\ \hline
\end{tabular}
\caption{ Comparison of AVA and UCF101-24}
\label{tab:datasets}
\end{table}

When partitioning AVA we had three goals: make it computationally tractable with our resources, maintain temporal continuity in the samples, and maintain the class distribution of the original dataset (see \cite{ava} for details). We ensured temporal continuity by sampling from the first segments from each split of the AVA dataset. We were also able to maintain a distribution that roughly follows Zipf's law, albeit with a smaller number of classes, as shown in Table \ref{table:miniAVA}. This ensured we would not simplify the dataset by reducing class imbalance inadvertently. Similarly to the testing scheme in the ActivityNet Task B challenge \cite{activitynetchallenge} involving AVA, we took particular care to make sure that all classes had at least 20 samples in the test set. We did this by removing the class labels that had less than 20 examples on the test set. The number of samples chosen for each set was 15000. We chose this number so that we could process the dataset in a reasonable time frame.

\subsection{Subsampling and Voting Scheme for AVA}
\begin{figure}[ht]
  \centering
  \includegraphics[width=.7\linewidth]{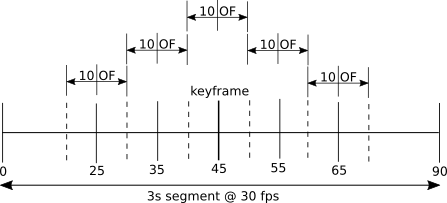}
  \caption{Our subsampling scheme. We extract 5 representative RGB frames and 5 OF stacks around a keyframe.}
  \label{fig:subsampling}
\end{figure}

As discussed in Sec. \ref{sec:ava}, the AVA dataset annotations are person-centric and made at a 1Hz frequency. Each annotation corresponds to a 3 second video segment. A straightforward implementation, at a rate of 30 FPS for each 3s segment, would need a receptive field of 90 frames and a 3D architecture to process a single segment. This would imply a much more computationally intense architecture and, as such, we propose a subsampling scheme as shown in Fig. \ref{fig:subsampling}.

Since we have 5 frames and OF volumes representative of a segment, we use a voting scheme to arrive at a consensus for the label to assign to a segment. A similar strategy was employed in \cite{ssn}.

As such, to obtain the predictions for a single segment we pass each frame of our 5 subsampled representative frames and its corresponding OF through the network and store the predictions. For the mutually exclusive (pose classes) predictions we count the maximum valued prediction as a vote and for all non-mutually exclusive (human-human and human-object classes) predictions we count any prediction values above a certain threshold $v$ as a valid vote. This threshold was initialized at 0.4, following \cite{ava}.
Then, for the mutually exclusive classes we take the most voted class as the predicted class for the segment and for each of the non mutually exclusive classes we take their top 3 most voted classes as the predicted classes for the segment. For the non-mutually exclusive classes the number of predicted classes will be between 0 (if no prediction is above the threshold) and 3. 

\begin{table}[ht]
\centering
\begin{tabular}{llll}
\hline
Class Category     & miniAVA & AVA & Mutually Exclusive     \\ \hline
Pose & 10   &   14 & Yes\\
Human-Object        & 12  & 49 & No\\
Human-Human        & 8  & 17 & No\\
\hline
\end{tabular}
\caption{Class categories in the AVA \cite{ava} dataset vs miniAVA.}
\label{table:miniAVA}
\end{table}

\subsection{Implementation Details}
Following the recommendations of \cite{DBLP:journals/corr/WangXW015} the training was done for 200 epochs. Batch size is 32 with a learning rate of 0.001 decaying to 0.0001 after 80\% of the epochs.
All streams were ResNet50 \cite{DBLP:journals/corr/HeZRS15} networks which were initialized with weights from \cite{feichtenhofer-resnet} for the respective RGB and OF networks trained on the UCF101 dataset.
The input to the networks are 224x224 RGB images and stacks of 10 OF volumes. 
For the results we use the same benchmarking tools as those provided for the AVA challenge with minor alterations only to obtain useful plots.

We report the results of several experiments on our miniAVA and on UCF101-24. This task involves localizing the atomic actions in space and time, achieving the highest video \ac{mAP} (at 0.5 \ac{IoU}) possible. For all experiments we round \ac{mAP} to 2 decimal places. 

\subsection{Tests}
We ran the following tests on our methods, on both the miniAVA dataset and the UCF101-24 datastet:
\begin{itemize}
    \item Baseline RGB, OF, and 2-stream Fusion with no attention Filtering
    \item Generalized Binary Loss vs Sum-of-Sigmoids.
    \item \ac{GBB}, Crop and Fovea attention filtering on RGB streams.
    %\item OF vs OF with Crop Filtering
    \item 2-stream Fusion With all combinations of attention filtering (including no filtering)
\end{itemize}

\section{Results}
\label{sec:results}
\subsection{Baseline}
For the baseline experiment, we want to establish what the mAP of an unaltered, state-of-the-art, Two-Stream \ac{CNN} is on this task, as well as the performance of each individual stream. With this baseline we have a result against which we can establish a fair comparison of the performance of our method.
\begin{table}[]
\centering
\begin{tabular}{c|c|c}
\hline
Model\textbackslash{}Dataset & AVA    & UCF101-24 \\ \hline
RGB                          & 5.06\% & 53.4\%    \\
\ac{OF}                         & \textbf{5.85}\% & 66.8\%    \\
RGB + \ac{OF}                   & 5.00\% & \textbf{73.3}\%    \\ \hline
\end{tabular}
\caption{Baseline individual streams and their fusion. Results are the video mAP at 0.5 IoU.}
\label{table:baseline}
\end{table}

In Table \ref{table:baseline} we show the results of this experiment.
For the AVA dataset we note how using only Optical Flow performs better than RGB. This has been also found to be the case for some implementations of these types of networks \cite{DBLP:journals/corr/CarreiraZ17} and is often due to the fact that certain actions have very clear motion patterns. Furthermore, we highlight that the baseline of the fusion is lower than both of the fused streams, which suggests that the spatiotemporal features being learned are not properly using complementary information from both streams, a result which we improve upon in further experiences.

On UCF101-24 the performance is a lot better for every model. This is another evidence of the high difficulty of the AVA dataset classification task. The results obtained in UCF101-24 are as not surprising, with a single \ac{OF} stream outperforming a single RGB stream. The Two-Stream fusion approach outperforms either single stream.

\subsection{Single Stream RGB Attention Filtering}
In this experiment, the goal is to compare the attention filtering approaches we tested: no-filtering, crop, \ac{GBB} and fovea attention filtering.
\label{subsec:singleRGB}
\begin{table*}[t]
\centering
\begin{tabular}{c|c|l|c|l}
\hline
Model\textbackslash Dataset & AVA    & \begin{tabular}[c]{@{}l@{}}Relative\\ Improvement\end{tabular} & UCF101-24                        & \begin{tabular}[c]{@{}l@{}}Relative\\ Improvement\end{tabular} \\ \hline
RGB                                           & 5.06\% & --                                                             & \textbf{53.4\%} & --                                                             \\ \hline
RGB + \ac{GBB}               & \textbf{5.63\%} & \textbf{+11.2\%}                                                        & 45.3\%                           & -15.1\%                                                        \\
RGB + Crop                                    & 5.19\% & +2.5\%                                                         & 24.5\%                           & -54.1\%                                                        \\
RGB + Fovea                                   & 5.12\% & +1.1\%                                                         & 49.2\%                           & -7.8\%                                                         \\ \hline
\end{tabular}
\caption{Attention filtering results on individual RGB streams vs baseline (RGB).Results are the video mAP at 0.5 IoU.}
\label{table:attentionrgb}
\end{table*}

In Table \ref{table:attentionrgb} we show the results of this experiment. For the AVA dataset, two main conclusions can be drawn from these results. One is that the use of all pre-filtering attention mechanisms improve results. The second is that the filtering techniques we hypothesize would lead to the networks learning artificial edges (i.e all except fovea filter) perform best. We think this is due to the fact that these artificial edges seem to be contributing to more accurate prediction of certain over represented classes, particularly \textit{stand}, which is the most common class. An important observation to make is that the fact that attention streams provide improvement on the AVA dataset is not surprising. In this dataset, seldom do the labels of all targets in a video coincide, leading to incorrect classifications when no attention is applied. If the network doesn't know \textit{who} it should classify, how can it be expected to correctly predict the label of two different targets in the same video when their labels are not the same? We believe this to be why our attention approach provides the biggest improvement in the AVA dataset. We build on this notion in subsequent tests.

On the UCF101-24 dataset the RGB stream with no attention filtering achieves the best results. This is not surprising taking into account the characteristics of the UCF101-24 dataset. Although the dataset is multi-person, in a single video, every person will be performing the same activity. As such, allowing the network to look at what a person that is not the target is doing is not detrimental to the networks performance. Regarding the attention, we have Fovea as the top performer, \ac{GBB} in second place and Crop in last. 

\subsection{Generalized Binary Loss}
In this experiment, we validate the choice of using our custom loss by showing the results of the best individual RGB stream when trained with a single output layer consisting only of sigmoid activation functions. This sum-of-sigmoids loss function is employed in the AVA baseline \cite{ava}.

\begin{table}[h]

    \centering
    \begin{tabular}{ll}
        Model\textbackslash{}Dataset &  AVA \\ \hline
        RGB + GBB - Generalized Binary Loss & \textbf{5.63\%} \\
        \hline
        RGB + GBB - Sum-of-Sigmoids & 5.01\% \\ 
        \hline
    \end{tabular}
    \caption{Results that validate our choice of a custom loss against a standard binary cross entropy loss.}
    \label{table:sigmoids}
\end{table}

In Table \ref{table:sigmoids} we show the results of this experiment. Our loss achieves an video mAP of $5.63\%$, representing a relative mAP improvement of $12.6\%$ over using a standard sum-of-sigmoids loss. We believe this is due to the fact that our Generalized Binary Loss takes into account the mutual exclusion of the pose classes, while the Sum-of-Sigmoids approach does not.%pose classes, stand is highly over represented and binary cross entropy is a sum of independent binary losses. In situations where mutual exclusivity could aid the network in learning to strongly predict a given pose class, the prediction might wrongly be attributed as stand.

% \subsection{Single Stream OF Attention Filtering}

% \begin{table}[h]
% \centering
% \begin{tabular}{ll}
% \hline
% Model\textbackslash{}Dataset    & AVA      \\ \hline
% \ac{OF}      & 5.85\%     \\
% \hline
% \ac{OF} + Crop & \textbf{5.90}\%  \\
% \hline
% \end{tabular}
% \caption{Attention filtering results on individual \ac{OF} streams vs baseline. Results are the mAP at 0.5 IoU.}
% \label{table:attentionflow}
% \end{table}

% In Table \ref{table:attentionflow} we show the results of this experiment. We only ran this experiment on the AVA dataset. We note that while we can see a very small improvement, that may also be due to certain artificial edges. However, the improvement of this method compared to the baseline is much smaller than in the RGB case, which leads us to believe that pre-filtering attention mechanisms for Optical Flow not only are not worthy to use but that their score is truly being extracted from motion features rather than introduced artificial edges.

\subsection{2-Stream Fusion Attention Filtering}
In this experiment we compare the results of the attention filters when fused in a two-stream \ac{CNN} architecture.
\label{subsec:2streamresults}

\begin{table*}[t]
\centering
\begin{tabular}{c|c|l|c|c}
\hline
Dataset                         & \multicolumn{3}{c|}{AVA}                                                            & UCF101-24 \\ \hline
Model                           & OF     & \begin{tabular}[c]{@{}l@{}}Relative\\ Improvement\end{tabular} & OF + Crop & OF        \\ \hline
RGB                             & 5.00\% & \multicolumn{1}{c|}{--}                                        & --        & 73.3\%    \\
RGB + \ac{GBB} & 3.59\% & -28.2\%                                                        & 4.16\%    & 70.9\%    \\
RGB + Crop                      & 5.01\% & +0.2\%                                                         & 5.06\%    & 68.7\%    \\
RGB + Fovea                     & \textbf{5.94\%} & \textbf{+18.8\%}                                                        & 4.95\%    & \textbf{74.5\%}    \\ \hline
\end{tabular}
\caption{Testing of several combinations of streams and their respective attention filters. Results are the video mAP at 0.5 IoU.}
\label{table:fusions}
\end{table*}

In Table \ref{table:fusions} we show the results of this experiment. For the AVA dataset, the first result is that cropped flow seems to worsen results when fused with all other streams except the cropped RGB stream, which is an interesting result that suggests some synergy in the learned features. We also note how the fovea filter performs better than all others when fused with unfiltered flow and that it is the only two-stream approach that improves on all previous experiments. %\textbf{We hypothesize that this might be due to the fact that the artificial edges introduced by the other filters would harm their performance when later merged with flow features. I would like to come up with an alternative explanation to this, but can't think of one at the moment.}

For the UCF dataset we again notice that the Crop filtering approach severely underperforming. This is to be expected, considering the poor performance this approach had on this dataset even in single-stream RGB attempts (See Section \ref{subsec:singleRGB}. However, a surprising result is found: the Two-Stream fusion with Fovea attention filtering is outperforming the no-filter approach. Like we stated in Section\ref{subsec:singleRGB}, we expected no-filter to be the best performer in the UCF101-24 dataset.

Our fovea attention filter outperforms all other approaches when fused with OF despite not being the best single stream approach (on UCF101-24 the best single stream approach was no-filter, and on AVA \ac{GBB} filtering. See Table \ref{table:attentionrgb}). This seems to indicate that the features learned using attention filtering on RGB better complement the features learned by the OF stream.

\section{Conclusions and Future Work}
\label{sec:conclusions}
%------------------------------------------------------------------------
In this work, we show our attention filtering approach to deal with multi-person, multi-label, spatiotemporal action detection datasets like the AVA dataset. We show improvements over the baseline on both datasets we used to run our experiments. Although we had to reduce the scale of our experiments due to computational limitations, we believe that our comparison of our method with and without attention filtering is reliable, and believe these results should be scalable to more complex architectures. We point out that the datasets we used for these experiments are new, and very challenging, particularly the AVA dataset, who was used in the ActivityNet challenge in CVPR 2018 (task B). Furthermore, we also present a comparison between our fovea attention filtering and other common approaches, like \ac{GBB} and crop filtering. We show that our fovea consistently outperforms those approaches, achieving a relative video mAP increase of $20\%$. We also achieve a relative video mAP improvement of $12.6\%$ when using our generalized binary loss over the standard sum-of-sigmoids.

In the future, we hope to study why fovea outperforms all other approaches when fused with an OF stream, even if this approach is not the best to use in single-stream RGB \ac{CNN}s. Furthermore, it would be very relevant to confirm that our approach works with different architectures. It would also be a natural extension of this work to test our approach on the full AVA dataset. Since the AVA baseline \cite{ava} is using a fusion of Faster RCNN \cite{DBLP:journals/corr/RenHG015} and I3D \cite{DBLP:journals/corr/CarreiraZ17}, the attention filtering they implicitly use is a crop on the feature maps after the region proposal network decides the RoIs. We would like to test what the results would look like if instead of cropping over the selected RoI a more sophisticated attention filter was employed, namely, our Fovea attention filtering that we show in this study to outperform cropping in two-stream networks (like the I3D architecture).

{\small
\bibliographystyle{ieee}
\bibliography{Bibliography}

\begin{thebibliography}{10}\itemsep=-1pt

\bibitem{rui}
A.~F. Almeida, R.~Figueiredo, A.~Bernardino, and J.~Santos-Victor.
\newblock Deep networks for human visual attention: A hybrid model using foveal
  vision.
\newblock In A.~Ollero, A.~Sanfeliu, L.~Montano, N.~Lau, and C.~Cardeira,
  editors, {\em ROBOT 2017: Third Iberian Robotics Conference}, pages 117--128,
  Cham, 2018. Springer International Publishing.

\bibitem{DBLP:journals/corr/abs-1803-01271}
S.~Bai, J.~Z. Kolter, and V.~Koltun.
\newblock An empirical evaluation of generic convolutional and recurrent
  networks for sequence modeling.
\newblock {\em CoRR}, abs/1803.01271, 2018.

\bibitem{laplacianpyramid}
P.~Burt and E.~Adelson.
\newblock The laplacian pyramid as a compact image code.
\newblock {\em IEEE Transactions on Communications}, 31(4):532--540, April
  1983.

\bibitem{DBLP:journals/corr/CarreiraZ17}
J.~Carreira and A.~Zisserman.
\newblock Quo vadis, action recognition? a new model and the kinetics dataset.
\newblock {\em 2017 IEEE Conference on Computer Vision and Pattern Recognition
  (CVPR)}, pages 4724--4733, 2017.

\bibitem{Csurka04visualcategorization}
G.~Csurka, C.~R. Dance, L.~Fan, J.~Willamowski, and C.~Bray.
\newblock Visual categorization with bags of keypoints.
\newblock In {\em In Workshop on Statistical Learning in Computer Vision,
  ECCV}, pages 1--22, 2004.

\bibitem{DBLP:journals/corr/DonahueHGRVSD14}
J.~Donahue, L.~A. Hendricks, M.~Rohrbach, S.~Venugopalan, S.~Guadarrama,
  K.~Saenko, and T.~Darrell.
\newblock Long-term recurrent convolutional networks for visual recognition and
  description.
\newblock {\em IEEE Trans. Pattern Anal. Mach. Intell.}, 39(4):677--691, Apr.
  2017.

\bibitem{feichtenhofer-resnet}
C.~Feichtenhofer, A.~Pinz, and R.~P. Wildes.
\newblock Spatiotemporal residual networks for video action recognition.
\newblock In {\em NIPS}, 2016.

\bibitem{feichtenhofer-multi}
C.~Feichtenhofer, A.~Pinz, and R.~P. Wildes.
\newblock Spatiotemporal multiplier networks for video action recognition.
\newblock In {\em 2017 IEEE Conference on Computer Vision and Pattern
  Recognition (CVPR)}, pages 7445--7454, July 2017.

\bibitem{feichtenhofer-fusion}
C.~Feichtenhofer, A.~Pinz, and A.~Zisserman.
\newblock Convolutional two-stream network fusion for video action recognition.
\newblock {\em CoRR}, abs/1604.06573, 2016.

\bibitem{activitynetchallenge}
B.~Ghanem, J.~C. Niebles, C.~Snoek, F.~Caba~Heilbron, H.~Alwassel, V.~Escorcia,
  R.~Khrisna, S.~Buch, and C.~Duc~Dao.
\newblock The activitynet large-scale activity recognition challenge 2018
  summary, 2018.

\bibitem{DBLP:journals/corr/GirdharRGSR17}
R.~Girdhar, D.~Ramanan, A.~Gupta, J.~Sivic, and B.~Russell.
\newblock {ActionVLAD: Learning spatio-temporal aggregation for action
  classification}.
\newblock In {\em {IEEE Conference on Computer Vision and Pattern
  Recognition}}, Honolulu, United States, 2017.
\newblock Project page: https://rohitgirdhar.github.io/ActionVLAD/.

\bibitem{deeplearning}
I.~J. Goodfellow, Y.~Bengio, and A.~Courville.
\newblock {\em Deep Learning}.
\newblock MIT Press, Cambridge, MA, USA, 2016.

\bibitem{ava}
C.~Gu, C.~Sun, D.~A. Ross, C.~Vondrick, C.~Pantofaru, Y.~Li,
  S.~Vijayanarasimhan, G.~Toderici, S.~Ricco, R.~Sukthankar, C.~Schmid, and
  J.~Malik.
\newblock Ava: A video dataset of spatio-temporally localized atomic visual
  actions.
\newblock In {\em The IEEE Conference on Computer Vision and Pattern
  Recognition (CVPR)}, June 2018.

\bibitem{DBLP:journals/corr/abs-1711-09577}
K.~Hara, H.~Kataoka, and Y.~Satoh.
\newblock Can spatiotemporal 3d cnns retrace the history of 2d cnns and
  imagenet?
\newblock {\em CoRR}, abs/1711.09577, 2017.

\bibitem{DBLP:journals/corr/HeZRS15}
K.~He, X.~Zhang, S.~Ren, and J.~Sun.
\newblock Deep residual learning for image recognition.
\newblock {\em 2016 IEEE Conference on Computer Vision and Pattern Recognition
  (CVPR)}, pages 770--778, 2016.

\bibitem{jhmdb}
H.~Jhuang, J.~Gall, S.~Zuffi, C.~Schmid, and M.~J. Black.
\newblock Towards understanding action recognition.
\newblock In {\em Proceedings of the IEEE international conference on computer
  vision}, pages 3192--3199, 2013.

\bibitem{6165309}
S.~Ji, W.~Xu, M.~Yang, and K.~Yu.
\newblock 3d convolutional neural networks for human action recognition.
\newblock {\em IEEE Transactions on Pattern Analysis and Machine Intelligence},
  35(1):221--231, Jan 2013.

\bibitem{5540039}
H.~Jégou, M.~Douze, C.~Schmid, and P.~Pérez.
\newblock Aggregating local descriptors into a compact image representation.
\newblock In {\em 2010 IEEE Computer Society Conference on Computer Vision and
  Pattern Recognition}, pages 3304--3311, June 2010.

\bibitem{karpathy}
A.~Karpathy, G.~Toderici, S.~Shetty, T.~Leung, R.~Sukthankar, and L.~Fei-Fei.
\newblock Large-scale video classification with convolutional neural networks.
\newblock In {\em CVPR}, 2014.

\bibitem{BMVC.25.76}
A.~V. Ken~Chatfield, Victor~Lempitsky and A.~Zisserman.
\newblock The devil is in the details: an evaluation of recent feature encoding
  methods.
\newblock In {\em Proceedings of the British Machine Vision Conference}, pages
  76.1--76.12. BMVA Press, 2011.
\newblock http://dx.doi.org/10.5244/C.25.76.

\bibitem{DBLP:journals/corr/NgHVVMT15}
J.~Y. Ng, M.~J. Hausknecht, S.~Vijayanarasimhan, O.~Vinyals, R.~Monga, and
  G.~Toderici.
\newblock Beyond short snippets: Deep networks for video classification.
\newblock {\em CoRR}, abs/1503.08909, 2015.

\bibitem{DBLP:journals/corr/abs-1712-05080}
P.~Nguyen, T.~Liu, G.~Prasad, and B.~Han.
\newblock Weakly supervised action localization by sparse temporal pooling
  network.
\newblock {\em CoRR}, abs/1712.05080, 2017.

\bibitem{fisherencoding}
F.~Perronnin, J.~S{\'a}nchez, and T.~Mensink.
\newblock Improving the fisher kernel for large-scale image classification.
\newblock In K.~Daniilidis, P.~Maragos, and N.~Paragios, editors, {\em Computer
  Vision -- ECCV 2010}, pages 143--156, Berlin, Heidelberg, 2010. Springer
  Berlin Heidelberg.

\bibitem{DBLP:journals/corr/RenHG015}
S.~Ren, K.~He, R.~Girshick, and J.~Sun.
\newblock Faster r-cnn: Towards real-time object detection with region proposal
  networks.
\newblock In C.~Cortes, N.~D. Lawrence, D.~D. Lee, M.~Sugiyama, and R.~Garnett,
  editors, {\em Advances in Neural Information Processing Systems 28}, pages
  91--99. Curran Associates, Inc., 2015.

\bibitem{Reynolds2009GaussianMM}
D.~A. Reynolds.
\newblock Gaussian mixture models.
\newblock In {\em Encyclopedia of Biometrics}, 2009.

\bibitem{Russakovsky2015}
O.~Russakovsky, J.~Deng, H.~Su, J.~Krause, S.~Satheesh, S.~Ma, Z.~Huang,
  A.~Karpathy, A.~Khosla, M.~Bernstein, A.~C. Berg, and L.~Fei-Fei.
\newblock Imagenet large scale visual recognition challenge.
\newblock {\em International Journal of Computer Vision}, 115(3):211--252, Dec
  2015.

\bibitem{simonyan}
K.~Simonyan and A.~Zisserman.
\newblock Two-stream convolutional networks for action recognition in videos.
\newblock In {\em Proceedings of the 27th International Conference on Neural
  Information Processing Systems - Volume 1}, NIPS'14, pages 568--576,
  Cambridge, MA, USA, 2014. MIT Press.

\bibitem{soomro2012ucf101}
K.~Soomro, A.~R. Zamir, and M.~Shah.
\newblock Ucf101: A dataset of 101 human actions classes from videos in the
  wild.
\newblock {\em arXiv preprint arXiv:1212.0402}, 2012.

\bibitem{DBLP:journals/corr/TranBFTP14}
D.~Tran, L.~D. Bourdev, R.~Fergus, L.~Torresani, and M.~Paluri.
\newblock {C3D:} generic features for video analysis.
\newblock {\em CoRR}, abs/1412.0767, 2014.

\bibitem{DBLP:journals/corr/abs-1711-11248}
D.~Tran, H.~Wang, L.~Torresani, J.~Ray, Y.~LeCun, and M.~Paluri.
\newblock A closer look at spatiotemporal convolutions for action recognition.
\newblock {\em CoRR}, abs/1711.11248, 2017.

\bibitem{tran2018closer}
D.~Tran, H.~Wang, L.~Torresani, J.~Ray, Y.~LeCun, and M.~Paluri.
\newblock A closer look at spatiotemporal convolutions for action recognition.
\newblock In {\em Proceedings of the IEEE Conference on Computer Vision and
  Pattern Recognition}, pages 6450--6459, 2018.

\bibitem{Traver2010ARO}
V.~J. Traver and A.~Bernardino.
\newblock A review of log-polar imaging for visual perception in robotics.
\newblock {\em Robotics and Autonomous Systems}, 58:378--398, 2010.

\bibitem{idt}
H.~Wang and C.~Schmid.
\newblock Action recognition with improved trajectories.
\newblock In {\em Proceedings of the 2013 IEEE International Conference on
  Computer Vision}, ICCV '13, pages 3551--3558, Washington, DC, USA, 2013. IEEE
  Computer Society.

\bibitem{DBLP:journals/corr/WangXW015}
L.~Wang, Y.~Xiong, Z.~Wang, and Y.~Qiao.
\newblock Towards good practices for very deep two-stream convnets.
\newblock {\em CoRR}, abs/1507.02159, 2015.

\bibitem{tsn}
L.~Wang, Y.~Xiong, Z.~Wang, Y.~Qiao, D.~Lin, X.~Tang, and L.~V. Gool.
\newblock Temporal segment networks for action recognition in videos.
\newblock {\em IEEE transactions on pattern analysis and machine intelligence},
  2018.

\bibitem{DBLP:journals/corr/abs-1712-04851}
S.~Xie, C.~Sun, J.~Huang, Z.~Tu, and K.~Murphy.
\newblock Rethinking spatiotemporal feature learning for video understanding.
\newblock {\em CoRR}, abs/1712.04851, 2017.

\bibitem{DBLP:journals/corr/XuDS17}
H.~Xu, A.~Das, and K.~Saenko.
\newblock {R-C3D:} region convolutional 3d network for temporal activity
  detection.
\newblock {\em CoRR}, abs/1703.07814, 2017.

\bibitem{ssn}
Y.~Zhao, Y.~Xiong, L.~Wang, Z.~Wu, X.~Tang, and D.~Lin.
\newblock Temporal action detection with structured segment networks.
\newblock {\em ICCV, Oct}, 2, 2017.

\bibitem{DBLP:journals/corr/ZhuLNH17a}
Y.~Zhu, Z.~Lan, S.~D. Newsam, and A.~G. Hauptmann.
\newblock Hidden two-stream convolutional networks for action recognition.
\newblock {\em CoRR}, abs/1704.00389, 2017.

\end{thebibliography}
}

\end{document}